\documentclass[a4paper,twoside]{article}

\usepackage{epsfig}
\usepackage{subfigure}
\usepackage{calc}
\usepackage{amssymb}
\usepackage{amstext}
\usepackage{amsmath}
\usepackage{amsthm}
\usepackage{multicol}
\usepackage{pslatex}
\usepackage{apalike}
\usepackage{color}
\usepackage{SCITEPRESS}     

\begin{document}

\title{Towards a tracking algorithm based on the clustering of spatio-temporal clouds of points}

\author{\authorname{Andrea Cavagna\sup{1}, Chiara Creato\sup{1}, Lorenzo Del Castello\sup{1}, Stefania Melillo\sup{1}, Leonardo Parisi\sup{1}\sup{2}, Massimiliano Viale\sup{1}}
\affiliation{\sup{1} Istituto Sistemi Complessi, Consiglio Nazionale delle Ricerche, UOS Sapienza, 00185 Rome, Italy.}
\affiliation{\sup{2} Dipartimento di Informatica Universita', Sapienza, 00198 Rome, Italy}
}


\abstract{The interest in $3D$ dynamical tracking is growing in fields such as robotics, biology and fluid dynamics. Recently, a major source of progress in $3D$ tracking has been the study of collective behaviour in biological systems, where the trajectories of individual animals moving within large and dense groups need to be reconstructed to understand the behavioural interaction rules. Experimental data in this field are generally noisy and at low spatial resolution, so that individuals appear as small featureless objects and trajectories must be retrieved by making use of epipolar information only. Moreover, optical occlusions often occur: in a multi-camera system one or more objects become indistinguishable in one view, potentially jeopardizing the conservation of identity over long-time trajectories. The most advanced $3D$ tracking algorithms overcome optical occlusions making use of set-cover techniques, which however have to solve NP-hard optimization problems. Moreover, current methods are not able to cope with occlusions arising from actual physical proximity of objects in $3D$ space. Here, we present a new method designed to work directly in $3D$ space and time, creating $(3D~+~1)$ clouds of points representing the full spatio-temporal evolution of the moving targets. We can then use a simple connected components labeling routine, which is linear in time, to solve optical occlusions, hence lowering from NP to P the complexity of the problem. Finally, we use normalized cut spectral clustering to tackle $3D$ physical proximity.}

\onecolumn \maketitle \normalsize \vfill

\section{\uppercase{Introduction}}
\label{sec:introduction}

In recent years the interest in $3D$ tracking has grown significantly, both in academic fields as turbulence \cite{ouellette2006quantitative}, biology \cite{dell2014automated}, and the social sciences \cite{moussaid2012traffic} and in industrial fields like robotics \cite{michel2007gpu}, surveillance \cite{hampapur2005smart}, and autonomous mobility \cite{ess2010object}. Advances in technology contributed to improve the tracking results in terms of quality of the retrieved trajectories, at the same time lowering the system requirements. These progresses allow today the automatic tracking of large groups of objects in a way that was prohibitive only a few years back.

A particularly energetic boost of the research into $3D$ tracking has come from the study of collective behaviour in biological systems, as bird flocks \cite{attanasi2015greta}, flying bats \cite{wu2011efficient}, insect swarms \cite{straw2010multi} \cite{puckett2014searching} \cite{cheng2015novel} and fish schools \cite{butail20103d} \cite{perez2014idtracker}. The aim in this field is to use experimental data about the actual trajectories of individual animals to infer the underlying interaction rules at the basis of collective motion \cite{giardina2008collective}. The crucial issue of the tracking algorithm is then to avoid identity switches and minimise fragmented trajectories, as this may result into a biased, or even wrong, understanding of the biological mechanisms. 

Data on collective animal behavior are characterized by frequent occlusions lasting up to tens of frames and by a low spatial resolution such that animals appear as objects without any recognizable feature. The latter fact rules out from the outset the use of any feature-based tracking method \cite{vacchetti2004combining}. Optical occlusions, on the other hand, arise when two or more targets get close in the $3D$ space or in the $2D$ space of one or more cameras. Individual targets are not distinguishable anymore and the identities of the occluded objects are  mixed for several frames. Occlusions introduce ambiguities which can result in fragmented trajectories (best case scenario) or identity switches (worse case scenario), depending on the tracking approach used. An effective tracking method for the study of collective behaviour must find a way to deal with them.

We can define two types of occlusions: i) `simple' $2D$ optical occlusions happen when two (or more) objects become closer than the optical resolution only in the camera space; in this case proximity is just an illusion of projection and the objects are {\it not} actually close in real $3D$ space, so that, in a multi-camera systems, there will always be one or more cameras in which the objects are well separated; ii) `hard' $3D$ occlusions occur when two (or more) objects get into actual physical proximity in real $3D$ space; in this case an optical occlusions is formed in {\it all} views of the multi-camera system. 

The most advanced tracking algorithms \cite{wu2011efficient} \cite{attanasi2015greta} \cite{cheng2015novel} successfully overcome the problem of $2D$ occlusions by using weighted set-cover techniques. This however requires solving a NP-hard optimization problem, with the consequent limitations on the maximum size of the studied system. On the other hand, even the most  robust tracking methods do not solve the problem of occlusions due to actual $3D$ proximity, hence incurring into switches of identity. We propose here a new 3D tracking algorithm -- name: Prometheus -- able to: i) solve occlusions due to $2D$ proximity making use of a polynomial time connected components labeling technique; ii) solve occlusions due to $3D$ proximity making use of a sophisticated clustering algorithm based on normalized cut methods.

\section{\uppercase{Related works}}

The first $3D$ tracking algorithms dealing with featureless objects were developed in the field of fluid dynamics, where the motion of passive tracer particles is studied to investigate  turbulent fluid flows. The most successful algorithm in this field is the one presented in \cite{ouellette2006quantitative}, which solves occlusion-related ambiguities locally in time, potentially producing fragmented trajectories. However, in the study of turbulence one can actually tune the density of tracers, so decreasing the optical density to a point where this is no longer critical. Clearly, this cannot be done in biological systems.

More recently the literature about $3D$ tracking on animal groups is growing, but the majority of the existing methods makes strongly use of objects' features and therefore they are not suitable for data coming from large, dense groups in the field. In this case, the requirement to have the whole group in the common field of view of all cameras implies a relatively low resolution at the individual level, making the targets quite featureless. To the best of our knowledge, the algorithms which best perform tracking of large natural systems are the ones in \cite{wu2011efficient} (bats), \cite{attanasi2015greta} (birds, insects) and \cite{cheng2015novel} (insects). These $3D$ tracking algorithms fist detect the objects moving in the common field of view of the camera system via standard background subtraction and segmentation. Links across cameras connect $2D$ objects, projections at the same instant of time of the same $3D$ target in different cameras. Links across time are then defined in the $2D$ space of each camera or in the $3D$ space. $2D$ objects are linked in time when representing the projection of the same target at subsequent frames time; while $3D$ reconstructed objects are linked when representing the three dimensional reconstruction of the same target at different instant of time. Depending on whether $2D$ or $3D$ links across time are used, $3D$ algorithm are classified as \textit{tracking-reconstruction} (TR) and \textit{reconstruction-tracking} (RT) algorithms \cite{wu2009tracking}. 

TR algorithms use $2D$ temporal links to create all the possible $2D$ paths in each camera. These $2D$ paths are then matched across cameras solving a weighted set-cover problems based on stereometric links. Conversely, RT algorithms use links across cameras to reconstruct  all  the existent $3D$ targets and then the correct trajectories are chosen following $3D$ temporal links. In both cases a global multi-linking approach is necessary to overcome occlusions, as any one-to-one local linking fails to recover the correct trajectories, producing highly fragmented and wrong tracks. The introduction of a global and multi-linking approach increases the computational complexity of the problem which has to be formulated as a NP-hard weighted set-cover. In the general case, the complexity of such a problem cannot be handled and only an approximation of the optimal solution can be found. In \cite{attanasi2015greta}, the set-cover problem is approached through a recursive technique, while in \cite{wu2011efficient} a greedy approximation is found. In \cite{cheng2015novel}, instead, the complexity of the problem is reduced choosing the trajectories globally in space but not in time. 

Neither TR nor RT methods can solve the occlusions due to $3D$ proximity described in the Introduction. Note that $3D$ occlusions  occur when the two objects become closer than the resolution of our $3D$ experimental setup, even though they do not literally occupy the same volume in  $3D$ space. For example, if our apparatus has an overall resolution (due to lens resolving power, atmospheric diffraction, sensor noise, etc) of $0.2$ meters, when two objects in a group become closer than this limit (which may happen due to inter-individual distance fluctuations), we have a $3D$ proximity occlusion. Hence, $3D$ proximity occlusions are more frequent than what one would naively expect. 

\begin{figure*}[!ht]
  \centering
   {\epsfig{file = 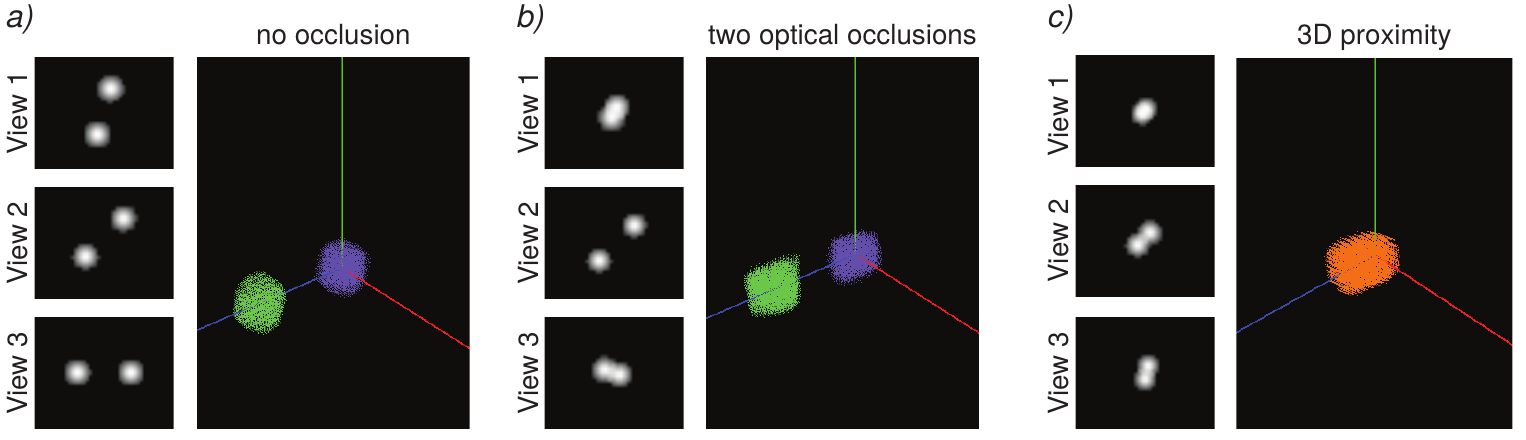, width = 15.5cm}}
  \caption{The $3D$ clouds created by Prometheus corresponding to: \textit{(a)} two targets well separated in all the views; \textit{(b)} two targets separated in the $3D$ space, but forming optical occlusions in two out of three cameras; \textit{(c)} two targets in actual $3D$ proximity (occlusion in all cameras). The two targets are reconstructed as two well-separated $3D$ clouds when they are not occluded in at least one camera, while they are reconstructed as a single $3D$ cloud when in $3D$ proximity. View 1, View 2 and View 3 show the image on each camera.}
  \label{fig1}
\end{figure*}

\section{\uppercase{Structure of the algorithm}}

The proposed algorithm (name: Prometheus) is a reconstruction-tracking method, since it first reconstructs targets in the $3D$ space and then it retrieves $3D$ trajectories. However, our method differs from classic reconstruction-tracking algorithms, as the one described in \cite{cheng2015novel}, because it does not work on the $2D$ barycenter of segmented objects and for this reason it does not need any cumbersome segmentation routine, but only a background subtraction. The algorithm can be broken into four steps: 1) background subtraction; 2) creation of the cloud of points; 3) Connected Components Labeling (CCL); 4) Normalized Cut Spectral Clustering (NCSC).

{\bf 1 - Background subtraction.} This is, of course, the most standard and by far least demanding part of the method. In order to discard background pixels, a background subtraction routine is performed, making use of a standard sliding window technique. This procedure is strengthened against image noise applying standard denoising and thresholding routines (see, for example, \cite{bgslibrarychapter} for a general description of background subtraction).

{\bf 2 - Creation of the cloud of points.} This module is the core of the new method. At each time frame Prometheus makes use of the geometric constraints of the camera system (stereometric and epipolar relations) to match pixels across cameras. In this way, for every set of pixels matched, it reconstructs the correspondent $3D$ point in the world space, hence creating a cloud of $3D$ points. Using three cameras, this process is performed defining triplets of pixels, one for each camera, that respect the trifocal constraint \cite{Hartley2004}. 

Let us illustrate what this procedure produces when we are dealing with two different targets (at fixed time). The easiest case is when the two targets are well separated in $3D$ space {\it and} they are not optically occluded in any camera view. In this case, of course, the method produces two separated cloud of $3D$ points (Fig.\ref{fig1}a). 

The second case is that of an optical occlusion, i.e. the targets are well separated in $3D$ space, but they form a single object in one (or more) of the $2D$ views (Fig.\ref{fig1}b -- two-cameras occlusion). This is what is normally hard to solve by tracking, as the $2D$ trajectories have an indecision point split at this instant of time. However, by working directly with the cloud of points in $3D$ space we see that the two objects become well-separated in $3D$, despite some slight shape deformation due to an epipolar echo of one object onto the other (Fig.\ref{fig1}b). 

\begin{figure}[!h]
  \centering
   {\epsfig{file = 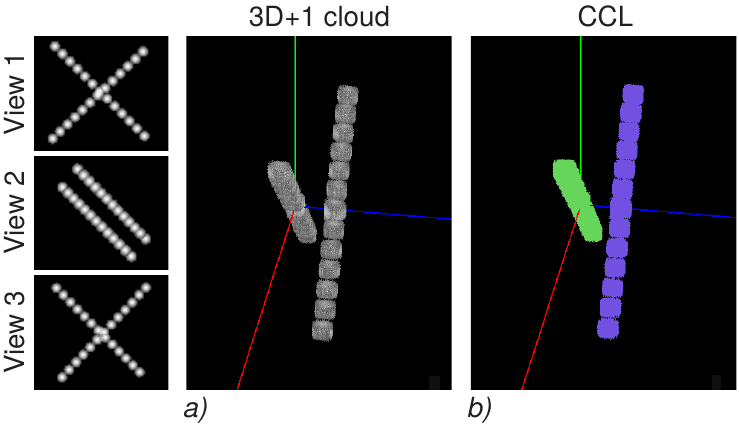, width = 7.5cm}}
  \caption{$2D$ occlusion. Left: the temporal evolution of two moving targets, seen by the three different cameras, forming an optical occlusion for a few frames in View 1 and View 3. \textit{(a)} the $(3D~+~1)$ clouds of points created by Prometheus. \textit{(b)} the two $(3D~+~1)$ clouds of points clustered by the CCL algorithm, identifying the two objects.}
  \label{fig2}
\end{figure}

\begin{figure*}[!ht]
  \centering
   {\epsfig{file = 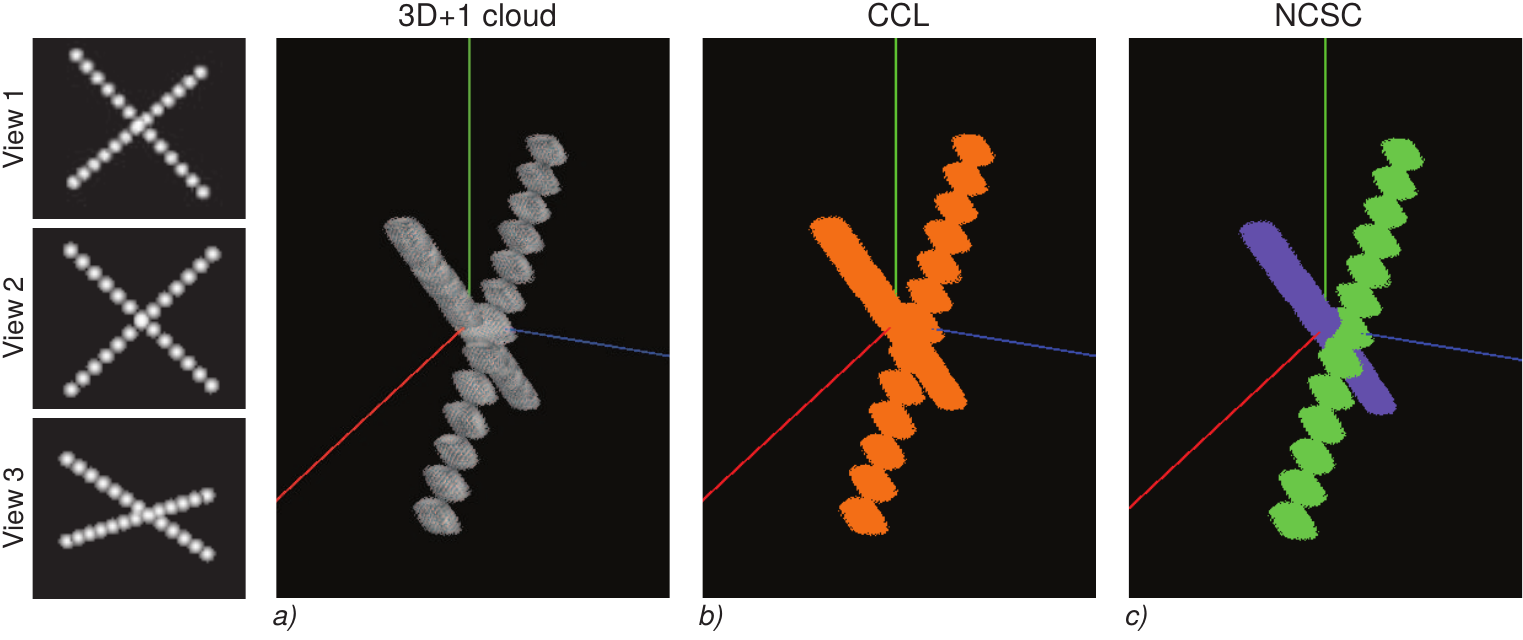, width = 13.0cm}}
  \caption{$3D$ proximity. Left: temporal evolution of two moving targets forming an optical occlusion in all the three cameras ($3D$ proximity). \textit{(a)} the $(3D~+~1)$ clouds of points created by Prometheus; \textit{(b)} the outcome of CCL clustering, clearly unable to retrieve the $3D+1$ volumes of the two objects, because they are in physical proximity for a few frames; \textit{(c)} the result of the NCSC clustering algorithm: the two objects are correctly clustered and identified.}
  \label{fig3}
\end{figure*}

When, on the other hand, the two targets are occluded in all three cameras (Fig.\ref{fig1}c), their correspondent volumes are no longer separated in the $3D$ space and indeed they become one single $3D$ cloud. As we have already said, in this case the two real objects in $3D$ space are closer than our resolution. 

Once all frames are processed, what we have is a global $(3D~+~1)$ cloud representing the volume of the full spatio-temporal evolution of the targets. As shown in Figs.\ref{fig2} and \ref{fig3}, the trajectory of each object appears as a spatio-temporal tube and the challenge now is to separate volumes corresponding to different targets. This is what we do by using clustering algorithms.

{\bf 3 - Connected Components Labeling (CCL).} The $(3D~+~1)$ cloud is partitioned in clusters separated in the $(3D~+~1)$ space. Such spatio-temporal clustering needs a notion of proximity to connect points in both space and time. Two $3D$ points belonging to the same frame are connected if their mutual distance is smaller than a fixed static threshold; two $3D$ points belonging to subsequent frames are connected when their mutual distance is smaller than a fixed dynamic threshold. The $(3D~+~1)$ cloud is now interpreted as a graph and it is clustered by using any Connected Components Labeling (CCL) technique \cite{Stockman2001}.

Fig.\ref{fig2} shows the situation represented by two moving objects never in physical $3D$ proximity, but occluded for a few frames in two of the three cameras. In this case, the CCL technique successfully separates the two identities, overcoming the optical occlusion. We stress that this is exactly the case that needs to be tackled by path multi-branching and set-cover techniques by other methods, which requires solving an NP-hard problem. Within Prometheus, on the other hand, this case is solved by CCL, which is merely P complex. This advantage may seem minor in the schematic case of \ref{fig2} (and it is), but it becomes a substantial aid when analyzing complex data, as that presented in the next Section.

When the two targets are in $3D$ proximity for a few frames (occlusion in all cameras Fig.\ref{fig3}), some $3D$ points of one target are linked in both space and time to  $3D$ points of the other target, connecting the two $3D$ volumes corresponding to the different targets, Fig\ref{fig3}a. Hence, in this case the CCL algorithm produces one single connected component and it fails to solve the occlusion, as shown in Fig.\ref{fig3}b. For this reason we need to resort to a more sophisticated clustering technique.

{\bf 4 - Normalized Cut Spectral Clustering (NCSC).} Consider the schematic case of Fig.\ref{fig3}, where two targets get in $3D$ proximity for a few frames. The two $3D$ volumes describing the spatio-temporal evolution of the two targets are connected by a few links concentrated around those few frames, while they are well separated in all the other frames. This suggests to use both spatial and temporal information to discard links connecting the two different targets, dividing the original cluster into two different connected components, each representing the dynamic evolution of one target. The cumulative weakness of the links connecting the different targets is the key motive to solve $3D$ proximity occlusions by spectral clustering.

\begin{figure*}[!ht]
  \centering
   {\epsfig{file = 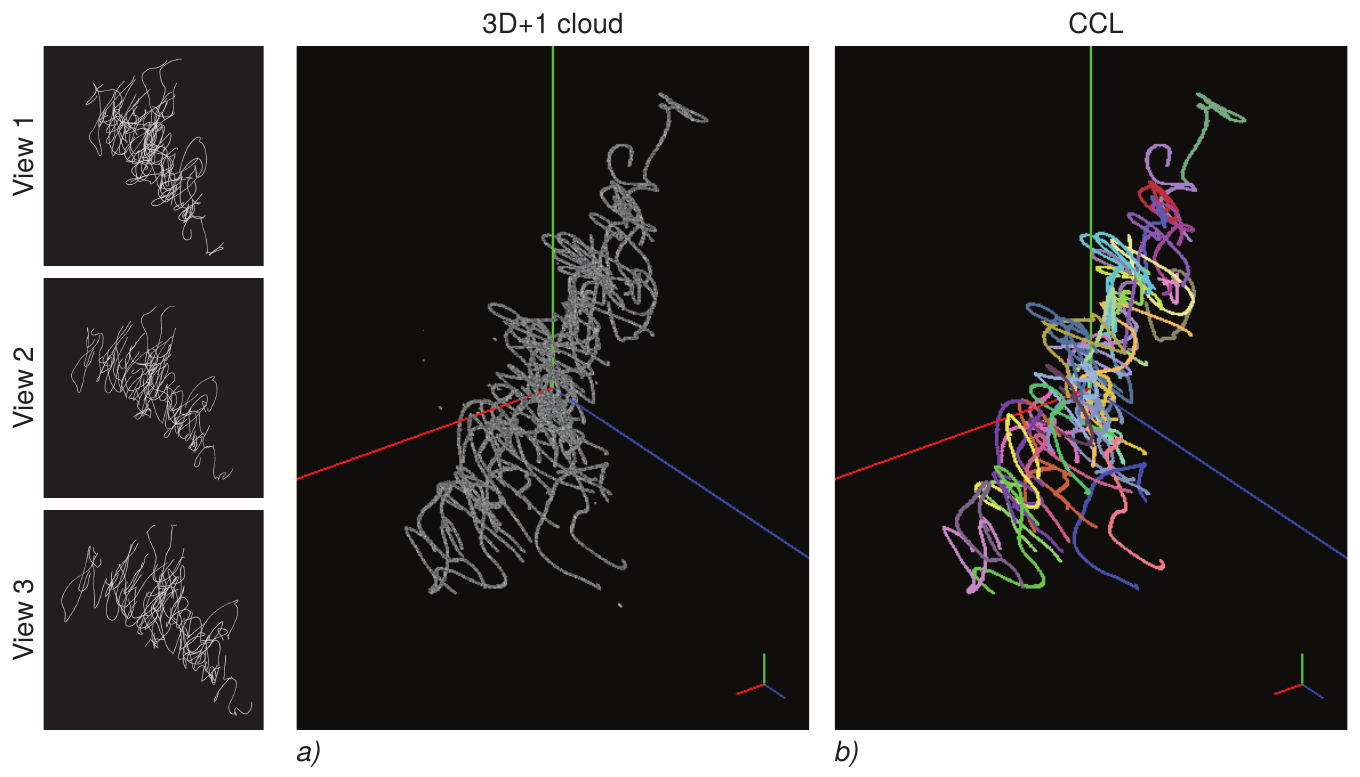, width = 13.0cm}}
  \caption{Test on semi-natural data. Left: temporal evolution of a swarm of $42$ midges for $200$ frames in the three different cameras; \textit{(a)} the $(3D~+~1)$ clouds of points; \textit{(b)} the result of Prometheus.}
  \label{fig4}
\end{figure*}

We work with a technique based on the Normalized Cut Spectral Clustering (NCSC) method introduced in \cite{shi2000normalized}. The NCSC approach is a substantial improvement on the Minimum Cut (MC) criterion \cite{Cormen2009}. MC defines the optimal partition of a graph as the one obtained by cutting across the minimum number of links; this favours the formation of small sets of isolated nodes in the graph. NCSC, on the other hand, optimizes the balance between cutting a small number of links {\it and} keeping the two clusters as even as possible in terms of points mass. This means that NCSC will try to minimize link-cutting while maximizing the equivalence in size of the output clusters. 

This spirit of NCSC seems very well suited to deal with the problem of splitting different trajectories (Fig.\ref{fig3}). In general, we want to track a group of targets all of similar size and shape (which is exactly the reason why we cannot perform the much-easier feature-based tracking), evolving for the same number of frames. For this reason, we expect that different targets occupy similar spatio-temporal volumes, so that NCSC, with its emphasis on creating balanced clusters, will divide them into the correct trajectories.

We apply NCSC to each multi-object cluster unsplit by CCL due to $3D$ proximity; we overcome the NP-complete complexity of NCSC by embedding it in the real values domain, thus finding a discrete approximation of the optimal solution in polynomial time \cite{shi2000normalized}. In this respect, we deal with hard $3D$ proximity occlusions similarly to what former methods deal with simple optical occlusions: we formulate the problem in terms of NP optimization, whose complexity is then tamed through a P approximated solution.

As shown in the schematic case of Fig.\ref{fig3}c, the normalized cut finds the two correct connected components and the two corresponding trajectories are thus correctly retrieved. It is perhaps interesting to note that in this way the concept of normalized cut is effectively extended from segmentation to tracking.

\section{\uppercase{Testing the algorithm}}

We performed tests of Prometheus on a semi-natural data set. We do this (instead of working directly on raw natural data) in order to have at the same time a biologically realistic data set and a ground truth with which comparing our results. Experimental data on midge swarms \cite{attanasi2014collective}, are tracked making use of the method of \cite{attanasi2015greta}); the resulting $3D$ trajectories are smoothed with $7$ points interpolation; moreover, the frame rate is doubled, passing from $170$ fps to $340$ fps linearly interpolating any pair of points. A system of three pinhole cameras is simulated and at each instant of time the $3D$ position of each target is projected on the three sensors (OpenGL library). Targets are monochromatic spheres of fixed radius, imaged as discs with Gaussian intensity profile. This procedure results in a set of images for each of the three cameras, which is given as an input to Prometheus, together with the trifocal tensor computed from the mutual positions of the three cameras \cite{Hartley2004}. 

Fig.\ref{fig4} shows the result of Prometheus on a semi-natural swarm of $42$ midges tracked for $200$ frames. Prometheus successfully solved all the $2D$ optical occlusions, producing $29$ isolated clusters corresponding to the insects which are never occluded in all the cameras at the same instant of time. We compared these trajectories with the ground truth and checked that they are correct. We remark that this result is achieved without the need of any set-cover technique.

Another $6$ clusters produced by CCL correspond to hybridised objects due to cases of $3D$ proximity, which should be tackled again by the NCSC routine. However occlusions in semi-natural datasets last longer than the ones the NCSC algorithm can currently solve, since some temporal constraints are still not implemented (see next Section).

\section{\uppercase{Future work}}

\begin{figure}[!h]
  \centering
   {\epsfig{file = 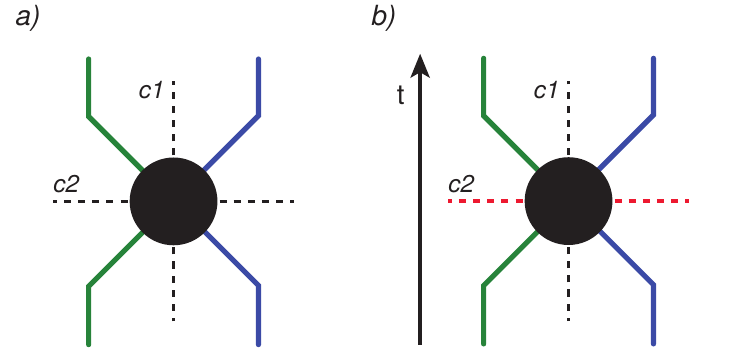, width = 7.5cm}}
  \caption{Long-term $3D$ proximity. Two $(3D~+~1)$ clouds well separated in time and space, represented in figure as the green and the blue lines, form an optical occlusion in all the cameras, due to long-term $3D$ proximity (black circle in figure). The dashed lines $c_1$ and $c_2$ depicts the two potential cuts evaluated by the NCSC algorithm. In absence of any bias differentiating time from space, the choice of the cut arbitrarily depends on the duration of the $3D$ proximity.}
  \label{fig5}
\end{figure}

Prometheus can solve $3D$ proximity occlusions lasting a few frames, but it currently does not handle long-term proximity problems. Fig.\ref{fig5} is a schematic representation of what may happen during a long-term $3D$ proximity occlusion. Two $(3D~+~1)$ clouds, represented as the green and the blue lines, are well separated except for those frames where the occlusion occurs; the resulting cloud of points is represented in the figure as the black circle. The NCSC routine has to find the partition which minimizes the weight of the discarded links while maximizing the similarity on the volumes occupied by the two resulting clusters. Depending on the time duration of the $3D$ proximity, it can either be more convenient for NCSC to cut along $c_1$ (correct choice) or along $c_2$ (wrong choice). 

This happens because, at its present state, the NCSC module of Prometheus does not differentiate between spatial and temporal dimension: cutting along time or space does not make any difference, provided that a right balance between links and mass is found. Of course, this does not need to be the case: time has a privileged status in the problem, so that {\it a priori} a cut longitudinal in time has to be preferred over one transverse in time. Hence, we plan to introduce a time bias in the NCSC linking and splitting algorithm to overcome this problem.

Secondly, proximity links based on a metric distance are suitable when the displacement of a single individual between two consecutive frames is smaller than the inter objects distance. This limitation can be overcome speeding up the frame rate. However, in practice this solution is not always feasible and we are planning  to introduce dynamic predictors in Prometheus in order to give more robust definition of temporal links.

\section{\uppercase{Conclusions}}

\label{sec:conclusion}

Current state-of-the-art tracking algorithms are able to overcome $2D$ optical occlusions formulating a NP-hard weighted set-cover problem, while they are not able to solve occlusions due to actual $3D$ proximity. We presented a new $3D$ tracking algorithm -- name: Prometheus -- that significantly improves this state of affairs.

Prometheus works directly in $3D$, retrieving the spatio-temporal volume occupied by each target in $(3D~+~1)$ dimensions. It solves $2D$ occlusions making use of a linear time connected components labeling routine, while it overcomes $3D$ proximity through a spectral clustering technique based on the NP-complete normalized cut. In this way, Prometheus makes NP-Complete what is currently considered impossible (actual $3D$ proximity), while making P what is currently NP-hard ($2D$ occlusions).

Preliminary tests on a semi-natural data set of insect swarms were performed to check the validity of the method. These tests confirmed our expectations, showing that the labeling technique together with the normalized cut approach is a promising new direction for a new generation of $3D$ tracking algorithm.

\section*{\uppercase{Acknowledgements}}

This work was supported by the following grants: IIT -- Seed Artswarm; ERC--StG No. 257126; and US-AFOSR No. FA95501010250 (through the University of Maryland). We thank Irene Giardina for discussions.

\bibliographystyle{apalike}

\vfill

\end{document}